\documentclass{article}
\usepackage{ijcai20}

\usepackage{times}

\usepackage{soul}
\usepackage{url}
\usepackage[utf8]{inputenc}
\usepackage[small]{caption}
\usepackage{graphicx}
\usepackage{amsmath}
\usepackage{booktabs}
\usepackage[T1]{fontenc}
\usepackage[ruled,lined]{algorithm2e}
\usepackage[bookmarks=false]{hyperref}

\newcommand{\tabincell}[2]{\begin{tabular}{@{}#1@{}}#2\end{tabular}}

\title{Dual Learning for Dialogue State Tracking}

\author{Zhi Chen, Lu Chen, Yanbin Zhao,Su Zhu and Kai Yu \\
  Key Lab. of Shanghai Education Commission for Intelligent Interaction and Cognitive Eng. \\
  SpeechLab, Department of Computer Science and Engineering \\
  Brain Science and Technology Research Center \\
  Shanghai Jiao Tong University, Shanghai, China \\
  {\tt \{zhenchi713,chenlusz,zhaoyb,paul2204,kai.yu\}@sjtu.edu.cn} }

\begin{document}

\maketitle

\begin{abstract}
In task-oriented multi-turn dialogue systems, dialogue state refers to a compact representation of the user goal in the context of dialogue history. Dialogue state tracking (DST) is to estimate the dialogue state at each turn. Due to the dependency on complicated dialogue history contexts, DST data annotation is more expensive than single-sentence language understanding, which makes the task more challenging. In this work, we formulate DST as a sequence generation problem and propose a novel dual-learning framework to make full use of unlabeled data. In the dual-learning framework, there are two agents: the primal tracker agent (utterance-to-state generator) and the dual utterance generator agent (state-to-utterance genera-tor). Compared with traditional supervised learning framework, dual learning can iteratively update both agents through the reconstruction error and reward signal respectively without labeled data. Reward sparsity problem is hard to solve in previous DST methods. In this work, the reformulation of DST as a sequence generation model effectively alleviates this problem. We call this primal tracker agent dual-DST. Experimental results on MultiWOZ2.1 dataset show that the proposed dual-DST works very well, especially when labelled data is limited. It achieves comparable performance to the system where labeled data is fully used.
\end{abstract}

\section{Introduction}
\label{sec:introduction}
Dialogue state tracker is a core part of the task-oriented dialogue system, which records the dialogue state. The dialogue state consists of a set of \emph{domain-slot-value} triples, where the specific value represents the user goal, e.g., $hotel(price=cheap)$. The dialogue system responds to the user just based on the dialogue state. Thus, in order to make the dialogue process natural and fluent, it is essential to extract the dialogue state from the dialogue context accurately. However, the paucity of annotated data is the main challenge in this field. In this work, we solve a key problem that how to learn from the unlabeled data in DST task. We design a dual learning framework for DST task, where the dialogue state tracker is the primal agent and the dual agent is the utterance generator. Within the dual learning framework, these two primal-dual agents help to update each other through external reward signals and reconstruction errors by using unlabeled data. It only needs a few of labeled dialogue data to warm up these two primal-dual agents.

However, there are two main challenges when combining dual learning framework with previous dialogue state tracking (DST) methods:

\textbf{How to represent dialogue state under dual learning framework?}
Dual learning method is first proposed in the neural machine translation (NMT) task. The outputs of the primal-dual agents in NMT task are both sequential natural languages. However, in DST task, the output of the dialogue state tracker consists of isolated domain-slot-value triples. The traditional DST task is formulated as a classification problem with the given ontology, where all the possible values of the corresponding slot are listed. Under this problem definition, the previous classification methods just choose the right value for each slot. The recent innovated tracker TRADE~\cite{wu-2019-transferable} directly generates the values slot by slot using copy mechanism from dialogue context. However, these tracker methods get slot values independently. During the dual learning loop, it is hard to get reward signal from these independent slot values. The reward signal from dual utterance generator is also hard to allocate to these isolated value generation processes. Since the relations of the predicted values are not modeled and they are assumed to be independent with each other, it would face serious reward sparse problem. In this work, we reformulate the dialogue state tracking task as a sequential generation task. The whole dialogue state is represented by a sequence with structured information. For example, the state $hotel(price=cheap, area=centre), taxi(destination=cambridge)$ can be represented as ``\textless$hotel$\textgreater $\ $ \textless$price$\textgreater $\ cheap$ \textless$area$\textgreater  $\ centre$ \textless$/hotel$\textgreater $\ $ \textless$taxi$\textgreater $\ $ \textless$destination$\textgreater $\ cambridge$ \textless$/taxi$\textgreater''.

\begin{figure*}[tb]
    \centering
    \includegraphics[width=0.8\textwidth]{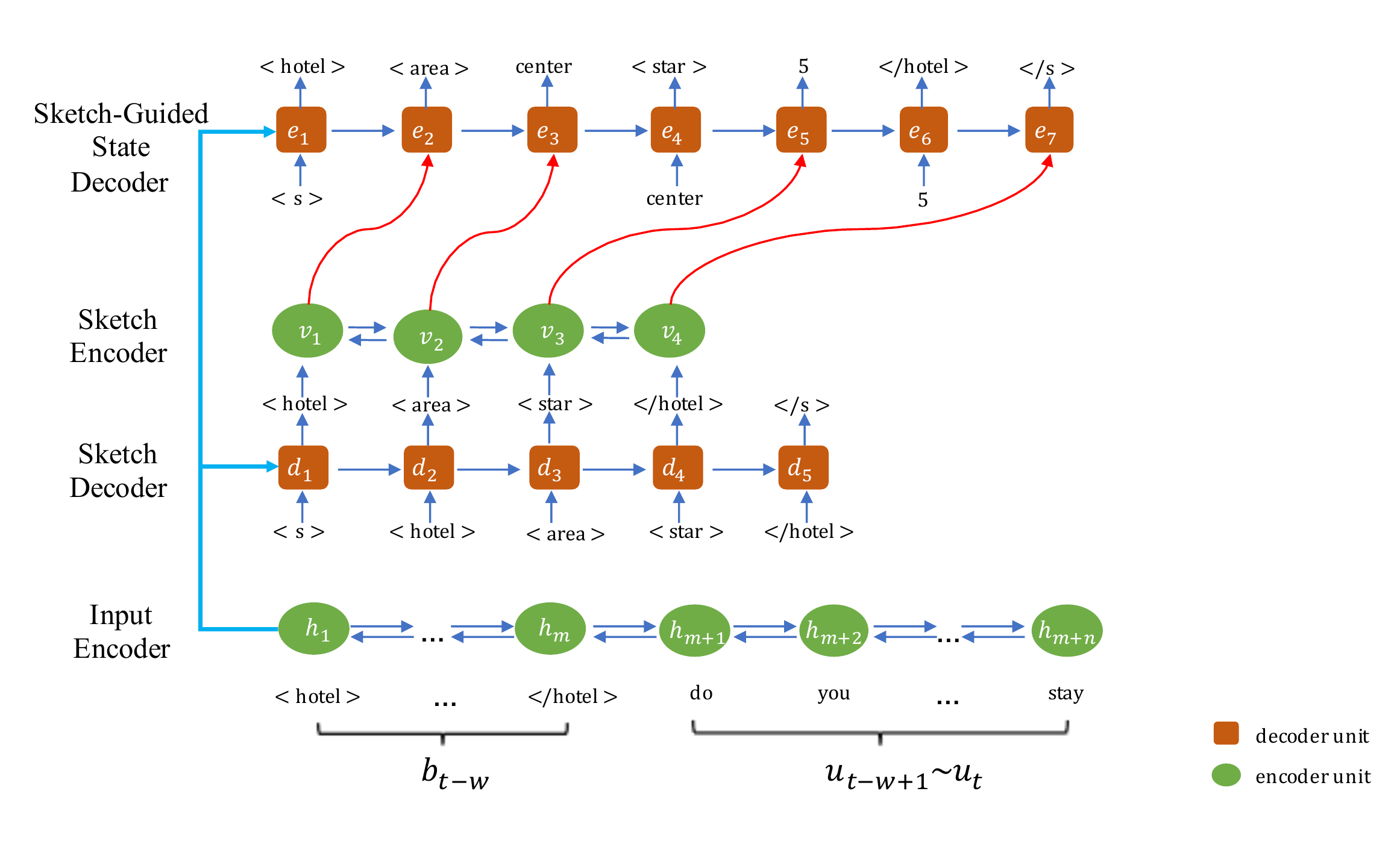}
    \caption{The coarse-to-fine tracker model, which consists of four parts: context encoder, state sketch decoder, sketch encoder and sketch-guided state decoder.}
    \label{fig:tracker}
\end{figure*}

\textbf{Is it reasonable that generating the whole dialogue context from dialogue state?}
The intuitive dual task of the state tracker is dialogue context generation. However, in MultiWOZ 2.1~\cite{eric2019multiwoz} dataset, the dialogue context has more than 10 turns on average and the average length of each sentence is over 10 tokens. It is very difficult in generating accurately a dialogue context with a dialogue state. Because the dialogue context is too long, it is hard to guarantee that the generated dialogue context contains the same semantics with the given state. In this work, we simplify the dual task into a user utterance generation task which ignores the specific values of the given state. The input of the dual task is composed of two parts (i.e., the delexicalized system utterance and the turn state), and its output is the delexicalized user utterance. The delexicalized script is copied from the released code \footnote{https://github.com/ConvLab/ConvLab}. The system utterance and user utterance can be lexicalized respectively according to the given turn state. We get a new pseudo-labeled dialogue turn. In order to produce multi-turn pseudo-labeled data, we sample a labeled dialogue data and combine it with the pseudo-labeled dialogue turn, where the dialogue turn directly adds to the end of the sampled dialogue context and the turn state covers into the label of the sampled state. Finally, we get a new dialogue context and pseudo label of the state, as the intuitive dual-task does.

The main contributions of this paper are summarized as follows:
\begin{itemize}
    \item An innovative dialogue state tracking framework based on dual learning is proposed, which can make full use of the unlabeled dialogue data for DST task.
    \item In this paper, we reformulate the dialogue state tracking as a sequence generation task and propose an efficient state generation model.
    \item In MultiWOZ 2.1 dataset, our proposed tracker achieves an encouraging joint accuracy. Under dual learning framework, when the labeled dialogue data is limited, the dual-DST works very well.
\end{itemize}

\section{Tracker and Dual Task}
In this section, we introduce the primal-dual models for DST task under dual learning framework. 

Different from previous DST approaches, we formulate the dialogue state tracking task as a sequence generation task. We represent the dialogue state as a structured sequence, rather than a set of isolated state triples. There are two important benefits: (1) The structured state representation keeps the relation information among the different slot values. The relation of these values contains some useful information, for example, the value of the slot \emph{departure} is different from the value of the \emph{destination} in flight ticket booking task. (2) Compared with isolated state representation, the state sequence is more applicable to the dual learning. It is easy to measure using BLEU score and evaluate using normal language model (LM)~\cite{mikolov2010recurrent}.

\subsection{Coarse-to-Fine State Tracker}
In this work, we adopt coarse-to-fine decoding method~\cite{dong2018coarse} to generate the sequential dialogue state. If specific values in sequential dialogue state are removed, we denote the rest representation as state sketch, which only contains domain-slot information, e.g., ``\textless$hotel$\textgreater $\ $ \textless$price$\textgreater $\ $ \textless$area$\textgreater $\ $ \textless$/hotel$\textgreater $\ $ \textless$taxi$\textgreater $\ $ \textless$destination$\textgreater $\ $ \textless$/taxi$\textgreater''. In order to simplify state generation, the coarse-to-fine method first generates the state sketch and then products the final state guided by the state sketch. The coarse-to-fine state generation model consists of four parts: dialogue context encoder, state sketch decoder, sketch encoder and sketch-guided state decoder, as shown in Fig.~\ref{fig:tracker}. 

\begin{figure*}[tb]
    \centering
    \includegraphics[width=\textwidth]{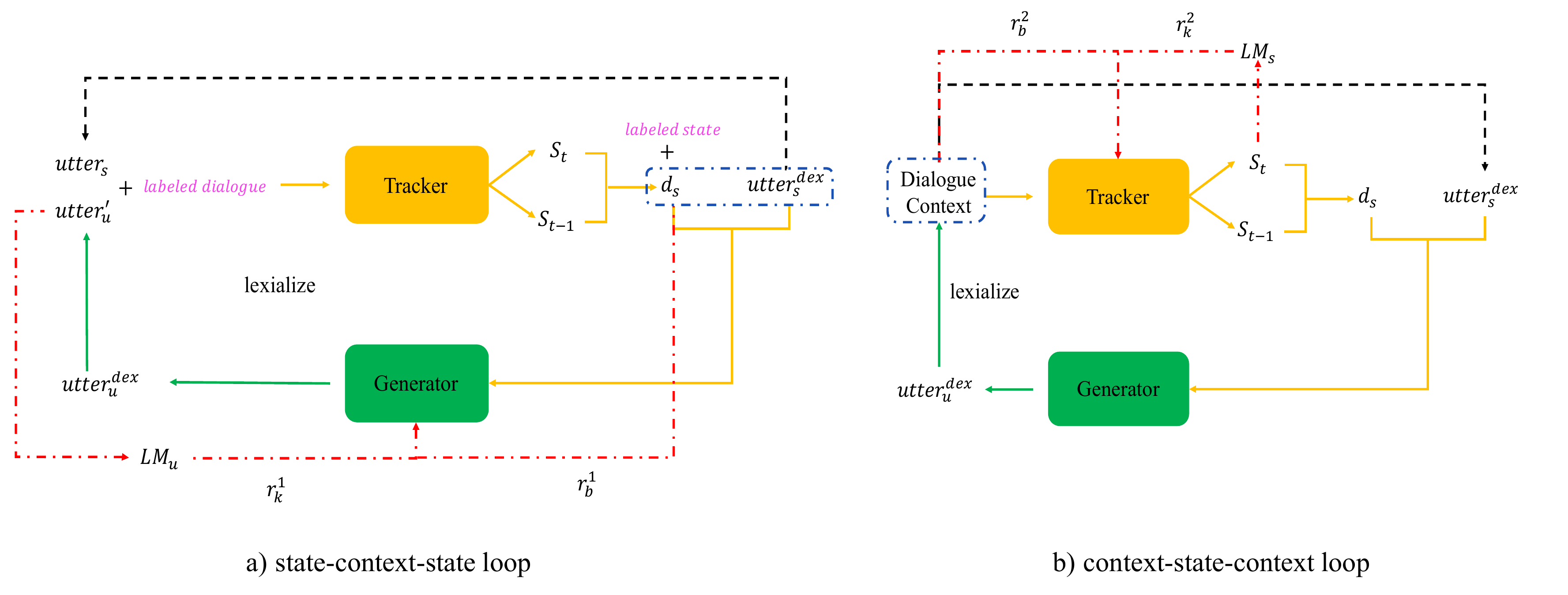}
    \caption{Abstraction of the dual learning framework. The dotted box means the start input content of the dual learning game.}
    \label{fig:abstract}
\end{figure*}

\begin{figure}[tb]
    \centering
    \includegraphics[width=\columnwidth]{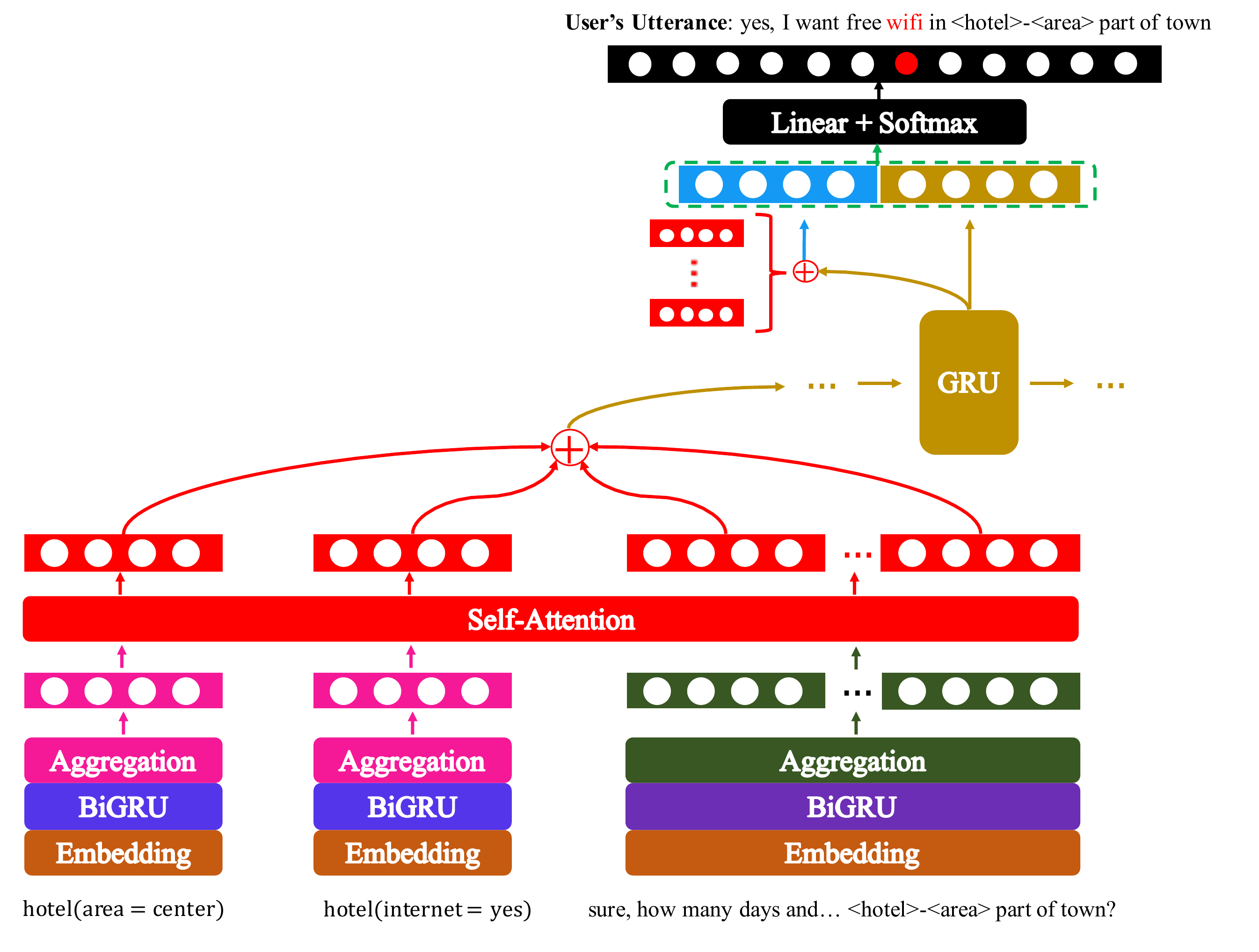}
    \caption{utterance generation model.}
    \label{fig:generator}
\end{figure}

\textbf{Context Encoder:} The input $x$ of coarse-to-fine tracker is composed of two components: current $w$ dialogue turns $u_{t-w+1}\sim u_t$ and $(t-w)$-th dialogue state $b_{t-w}$, where $w$ is window size of the dialogue context and earlier dialogue utterances is replaced by $(t-w)$-th dialogue state. In this work, we directly concatenate them together and use bi-directional gated recurrent units (GRU) to encode the input as:
\begin{align}
\overrightarrow{\mathbf{h}}_i &= f^{x}_{\text{GRU}}(\mathbf{h}_{i-1}, \mathbf{x}_i), i=1,\dots, |x|, \\
\overleftarrow{\mathbf{h}}_i &= f^{x}_{\text{GRU}}(\mathbf{h}_{i+1}, \mathbf{x}_i), i=|x|,\dots, 1, \\
\mathbf{h}_i &= [\overrightarrow{\mathbf{h}}_i, \overleftarrow{\mathbf{h}}_i]
\end{align}
where $\mathbf{x}_i$ is embedding of $i$-th token of the input $x$, $[\cdot, \cdot]$ means the concatenation of two vectors and $f^{x}_{\text{GRU}}$ is the input GRU function.

\textbf{State Sketch Decoder:} The sketch decoder generates a state sketch $a$ conditioned on the encoded context. We use a unidirectional GRU to decode the state sketch with the attention mechanism~\cite{luong2015effective}. At $t$-th time step of sketch decoding, the hidden vector is computed by $\mathbf{d}_t=f^{a}_{\text{GRU}}(\mathbf{d}_{t-1}, \mathbf{a}_{t-1})$, where $f^{a}_{\text{GRU}}$ is the GRU function and $\mathbf{a}_{t-1}$ is the embedding of previously predicted token. The initial hidden state $\mathbf{d}_0$ is $\overleftarrow{\mathbf{h}}_1$. The attention weight from $t$-th decoding vector to $i$-th vector in encoder is $s_{i}^{t}=\frac{\text{exp}(u_{i}^{t})}{\sum_{j=1}^{|x|} \text{exp}(u_{j}^{t})}$. The attention score $u_{i}^{t}$ is computed by
\begin{align}
u_{i}^{t} = \mathbf{v}^{T}\text{tanh}(\mathbf{W}_1 \mathbf{d}_t + \mathbf{W}_2 \mathbf{h}_i + \mathbf{b}),
\label{eq:attw}
\end{align}
where $\mathbf{v}$, $\mathbf{W}_1$, $\mathbf{W}_2$ and $\mathbf{b}$ are parameters. Then we calculate the distribution of the $t$-th sketch token $p(a_t|a_{<t})$ using
\begin{align}
p(a_t|a_{<t}) &= \text{softmax}(\mathbf{W}_a[\mathbf{d}_t, \mathbf{s}_t] +\mathbf{b}_a),  \\
\mathbf{s}_t &= \sum_{i=1}^{|x|}u_{i}^{t}\mathbf{h}_i,
\label{eq:prob}
\end{align}
where $\mathbf{W}_a$ and $\mathbf{b}_a$ are trainable parameters. Generation terminates until the end token of sequence ``$<$EOB$>$'' is emitted.

\textbf{Sketch Encoder:} We use another bidirectional GRU to map the sketch state into a sequence of sketch vectors $\{\mathbf{v}_i\}_{i=1}^{|a|}$, as context encoder does.

\textbf{Sketch-Guided State Decoder:} The final state generation is similar to sketch generation. The difference comes from that the state generation tries to use the generated sketch state. In sketch generation process, the input of the sketch decoder is always previously predicted token. However, during state generation, the input of state decoder at $t$-th time step $\mathbf{z}_t$ is
\begin{equation}
\label{eq:final_input}
\mathbf{z}_t=\left\{
\begin{aligned}
\mathbf{v}_k, \ \  \ \ &\text{$y_{t-1}$ is equal to $a_k$} \\
\mathbf{y}_{t-1}, \ \ \ \  &\text{otherwise},
\end{aligned}
\right.
\end{equation}
where $\mathbf{y}_{t-1}$ is the embedding of the predicted token at $(t-1)$-th time step.

\begin{algorithm}
\SetAlgoLined
\KwIn{unlabeled dialogue data $D_u$, unlabeled turn state $D_s$, labeled  dialogue-state pairs $(\hat{D}_u, \hat{D}_s)$, corresponding delexicalized dialogue context $\hat{D}_u^{\text{dex}}$, the language model of user utterance $LM_u$, the language model of coarse state $LM_s$, state tracker $P(\cdot|\Theta_{u2s})$, utterance generator $P(\cdot|\Theta_{s2u})$}
\BlankLine
 \Repeat{Convergence}{
  $\qquad$ $\qquad$ $\qquad$ $\qquad$ $\triangleleft$ State-Context-State Loop\;
  \BlankLine
  (1) Sample an unlabeled turn state $d_s$ from $D_s$ and a related delexicalized system utterance $uttr_s^{\text{dex}}$\;
  (2) Generate delexicalized user utterance $uttr_u^{\text{dex}}$ using generator $P(\cdot|\Theta_{s2u})$\;
  (3) Lexicalize $uttr_s^{\text{dex}}$ and $uttr_u^{\text{dex}}$ using turn state and get a dialogue turn $d_u =$ ($uttr_s$, $uttr_u$)\;
  (4) Evaluate the user utterance $uttr_u$ using $LM_u$ and get external-knowledge reward $r_k^1$ \;
  (5) Sample a labeled dialogue-state pair $(\hat{d}_u, \hat{d}_s)$ and combine this pair with $(d_u, d_s)$ to get a new dialogue-state pair $(\bar{d}_u, \bar{d}_s)$ \;
  (6) Update tracker $P(\cdot|\Theta_{u2s})$ using $(\bar{d}_u, \bar{d}_s)$\;
  (7) Generate the dialogue state $\bar{d}_s^{\prime}$ using tracker $P(\cdot|\Theta_{u2s})$ and get BLEU score reward $r_b^1$ with $\bar{d}_s$ \;
  (8) Update generator $P(\cdot|\Theta_{s2u})$ by policy gradient loss with reward $r^1=\alpha r_k^1 + (1-\alpha) r_b^1$\;
  \BlankLine
  $\qquad$ $\qquad$ $\qquad$ $\quad$ $\triangleleft$ Context-State-Context
  Loop\;
  \BlankLine
  (9) Sample a unlabeled dialogue context $d_u$ from $D_u$\;
  (10) Generate the dialogue state $s_t$ and previous dialogue state $s_{t-1}$ and get $t$-th turn state $d_s$\;
  (11) Evaluate the sketch of state $s_t$ using $LM_s$ and get external-knowledge reward $r_k^2$\;
  (12) Get delexicalized utterances ($uttr_s^{\text{dex}}$, $uttr_u^{\text{dex}}$) of $t$-th turn in $d_u$ with turn state $d_s$\;
  (13) Update generator $P(\cdot|\Theta_{s2u})$ by cross-entropy loss with $uttr_s^{\text{dex}}$, $d_s$ and $uttr_u^{\text{dex}}$\;
  (14) Generate the user utterance using $P(\cdot|\Theta_{s2u})$ with $uttr_s^{\text{dex}}$ and $d_s$ and lexicalize it into $uttr_u^{\prime}$\;
  (15) Calculate the BLEU score of $uttr_u^{\prime}$ as the reward $r_b^2$ and update tracker $P(\cdot|\Theta_{u2s})$ by policy gradient loss with reward $r^2=\alpha r_k^2 + (1-\alpha) r_b^2$\;
  \BlankLine
 }
 \caption{Dual learning method for dialogue state tracking}
 \label{alg:algorithm}
\end{algorithm}

\subsection{Dual Task}
As introduced in Section~\ref{sec:introduction}, the dual task of dialogue state tracker is simplified into a user utterance simulation task. 

\textbf{Encoder:} The input of utterance generation model is composed of two parts: turn state and system utterance (or wizard utterance). The turn state means the dialogue state mentioned by current dialogue turn, which consists of several domain-slot-value triples. We use a bidirectional GRU to encode each triple into a state vector respectively, as shown in Fig~\ref{fig:generator}. We map the system utterance into a sequence of token vectors. Then we use a self-attention layer~\cite{vaswani2017attention} to encode state vectors and token vectors together to get final encoded vector.

\textbf{Decoder:} The utterance decoder generates the user utterance conditioned on the designed turn state and system utterance. We use a unidirectional GRU to generate the user utterance with attention mechanism. The initial hidden state of the decoder is sum pooling of final encoded vector.

In the dual task, the given system utterance and the generated user utterance are delexicalized, which means that specific values of the dialogue state in two utterances are removed and replaced by common $domain$-$slot$ flags. For example, if the turn state is $hotel(star=5)$, the system utterance could be ``Do you want to reserve $<$hotel$>$-$<$star$>$ star hotel?'' and the user utterance could be ``Yes, I need $<$hotel$>$-$<$star$>$ star.''. Inversely, when the delexicalized utterance is given, we can use the corresponding turn state to get lexicalized utterance. Because the delexicalized system utterance is easy to collect, the function of the dual model can be regarded as to generate a lexicalized dialogue turn given a turn dialogue state.

\section{Dual Learning for DST}
In this section, we present the dual learning mechanism for dialogue state tracking. Before introduce the dual learning method for DST, we define the state tracking model and dual generation model as $P(\cdot|\Theta_{u2s})$ and $P(\cdot|\Theta_{s2u})$, respectively. Similar to dual-NMT~\cite{he2016dual}, we have also two pretrained language models to evaluate the generated state and user utterance, which are indicated as $LM_s$ and $LM_u$. Noticing that we pretrain language model for the sketch of the dialogue state, where the slot values are removed. We regard two language models as two kinds of external knowledge. The dual game of DST task consists of two sub-games: state reconstruction and utterance reconstruction. In other words, the dual learning method contains two kinds of training loop. The abstract of the dual learning method shows in Fig~\ref{fig:abstract}.

The first training loop for \textbf{state reconstruction} starts from a turn state. The utterance generator $P(\cdot|\Theta_{s2u})$ generates the delexicalized user utterance with a sampled delexicalized system utterance. Noticing that the sampled utterance normally contains some domain-slots. The generated utterance can be evaluated by $LM_u$. We use logarithmic of the utterance probability calculated by language model as external-knowledge reward $r_k^1$. Then we pair the given state and the generated utterances as pseudo labeled data to update the tracker $P(\cdot|\Theta_{u2s})$. Because this pair data contains only one turn, we sample from labeled multi-turn data and combine them together to get new multi-turn data. The tracker $P(\cdot|\Theta_{u2s})$ can further predict the state of concatenated utterances in the new multi-turn data. Then we can get BLEU score of the predicted state with combined state. The BLEU score can be regarded as another reward $r_b^1$ to indicate the quality of the generated utterance. At the end of this loop, the generator $P(\cdot|\Theta_{s2u})$ can be updated using weight-sum reward $r^1=\alpha r_k^1 + (1-\alpha) r_b^1$ by policy gradient loss~\cite{sutton2000policy}, where $\alpha$ is the hyper-parameter. The data flow of state reconstruction game is state-context-state.

The second training loop for \textbf{utterance reconstruction} starts from dialogue context with $t$ dialogue turns. The tracker $P(\cdot|\Theta_{u2s})$ predicts the $t$-th dialogue state $s_t$ and the previous state $s_{t-1}$. The sketch of the predicted state $s_t$ can be evaluated by $LM_s$. The external-knowledge reward $r_k^2$ is still logarithmic of the probability of the generated state sketch. Then we can get the $t$-th turn state $d_s$. We can further get $t$-th delexicalized system utterance $utter_s^{\text{dex}}$ using the $d_s$. The generator generates the user utterance with turn state $d_s$ and system utterance $utter_s^{\text{dex}}$. Then we calculate the BLEU score of the user utterance $utter_u^\prime$, which is lexicalized from the generated user utterance. The BLEU score is an implicit reward $r_b^2$ to measure the generated state. Similarly, the tracker $P(\cdot|\Theta_{u2s})$ can be updated using weight-sum reward $r^2=\alpha r_k^2 + (1-\alpha) r_b^2$ by policy gradient loss. The data flow of utterance reconstruction game is context-state-context. 

The specific process of the dual learning for DST is shown in Algorithm~\ref{alg:algorithm}, where state-context-state loop means state reconstruction process and context-state-context loop indicates utterance reconstruction process.

\section{Experiments}
\subsection{Dataset}
\label{sec:dataset}
We evaluate our methods in MultiWOZ 2.1 dataset, which is the largest task-oriented dialogue dataset for multi-domain dialogue state tracking task. MultiWOZ 2.1 dataset contains 8438 multi-domain dialogues and spans 7 dialogue domains. For dialogue state tracking task, there are only 5 domains (\emph{restaurant, hotel, attraction, taxi, train}) in validation and test set. The domains \emph{hospital, bus} only exist in training set. Around 70\% dialogues have more than 10 turns and the average length of the utterances in the dialogue is over 10.

\subsection{Training Details}
Similar to TRADE, we initialize all the embeddings using the concatenation of Glove embeddings~\cite{pennington2014glove} and character embeddings~\cite{hashimoto2017joint}. We set the window size as 10 turns. The hidden size of all GRUs is 500. Under the dual learning framework, there are two training phases: pretraining phase and dual learning phase. The pretraining phase aims to warm up the state tracker and the utterance generator with labeled data. We adopt Adam~\cite{kingma2014adam} optimizer with learning rate 1e-4. During the dual learning phase, the learning rate is 1e-5. In order to stabilize the dual learning, we still use the cross-entropy loss to update the above two models with labeled data. The reward weight $\alpha$ is 0.5.

\begin{small}
\begin{table}[h]
\begin{center}
\begin{tabular}{ c|c|c|c}
 \hline
Model & +BERT & \tabincell{c}{Joint Acc.\\MultiWOZ 2.1}  & ITC \\
 \hline
DS-DST & Y & 51.21\%   &   O(M) \\
SOM-DST & Y & 52.57\%   &   O(1)\\
DST-picklist & Y & \bf{53.30\%}   &   O(MN) \\
 \hline
$\text{HJST}$ & N  & 35.55\%   &   O(M) \\
$\text{DST Reader}$ & N  & 36.40\%  &   O(M) \\
$\text{FJST}$ & N  & 38.00\%  &   O(M)\\
$\text{HyST}$  & N & 38.10\%  &   O(M) \\
TRADE & N & 45.60\%  &   O(M) \\
Coarse2Fine DST(ours) & N & 48.79\%  &   O(1)\\
dual-DST(ours) & N & \bf{49.88\%}  &   O(1) \\
 \hline
\end{tabular}
\caption{The results of baseline models and our proposed coarse-to-fine tracker in MultiWOZ 2.1 dataset. +BERT means that the tracking model encodes the utterances using pretrained BERT. ITC means the inference time complexity, which measures the calculation time of evaluating state. In ITC column, $M$ is the number of slots and $N$ is the number of values. Joint Acc. means the joint goal accuracy.}
\label{tab:result}
\end{center}
\end{table}
\end{small}

\subsection{Baseline Methods}
We first compare our proposed coarse-to-fine state tracker with previous state tracking methods, when all the labeled training data is used.

\begin{itemize}
    \item \textbf{FJST}~\cite{eric2019multiwoz} and \textbf{HJST} ~\cite{eric2019multiwoz} are two straightforward methods, which directly predict all the slot values based on the encoded dialogue history. Instead of directly concatenating the whole dialogue history as input in FJST, HJST takes the hierarchical model as the encoder.
    \item \textbf{HyST}~\cite{goel2019hyst} is a hybrid method that improves HJST by adding the value-copy mechanism.
    \item \textbf{TRADE}~\cite{wu-2019-transferable} directly generates the slot value from the dialogue history.
    \item \textbf{DS-DST}~\cite{zhang2019find} and \textbf{DST-picklist}~\cite{zhang2019find} divide the slots as uncountable type and countable type and generate the slot value in a hybrid method like HyST. Compared with DS-DST, DST-picklist knows all the candidate values of the slots, including uncountable slots.
    \item \textbf{SOM-DST}~\cite{kim2019efficient} feeds dialogue history and previous state as the input and modifies the state with dialogue history into the current state.
    \item \textbf{DST Reader}~\cite{gao2019dialog} formulates DST task as a machine reading task and leverages the corresponding method to solve the multi-domain task.
\end{itemize}

The second experiment is to invalid the dual learning framework for DST task. In this experiment setup, we randomly sample 20\%, 40\%, 60\% and 80\% labeled data in training data. The rest data is used as unlabeled data. We compare dual learning method with \textbf{pseudo labeling method}, which is an important approach to use the unlabeled data. The pseudo labeling method first uses the sampled labeled data to pretrain our proposed tracker. During the training of pseudo labeling method, the pretrained tracker is used to generate the state of the unlabeled dialogue context. Then, the dialogue context and the generated state are paired together as the pseudo-labeled data to retrain the tracker. In order to stabilize the training process of the pseudo labeling method, we also mixture the pseudo labeled data and labeled data as a batch to update the pretrained tracker.

\subsection{Results}
\paragraph{The performance of our trackers:}
As shown in Table~\ref{tab:result}, our proposed coarse-to-fine tracker achieves the highest joint goal accuracy in the BERT-free models. Our proposed tracker directly generates all the slot values, which is represented as a structured sequence. Compared with the methods that predict the values slot by slot, the inference time complexity (ITC) is O(1). This property is important for the dialogue system. The response time of the dialogue system effects user experience seriously. Compared with SOM-DST, our proposed tracker does not rely on the pretrained BERT~\cite{devlin2018bert}, whose model size is more than 110M. This is another challenge for memory-starve devices. Compared with recently proposed TRADE, our proposed coarse-to-fine tracker not only reduces the inference time, but also gets the absolute 3.19\% joint goal accuracy improvement. 

\begin{figure}[tb]
    \centering
    \includegraphics[width=\columnwidth]{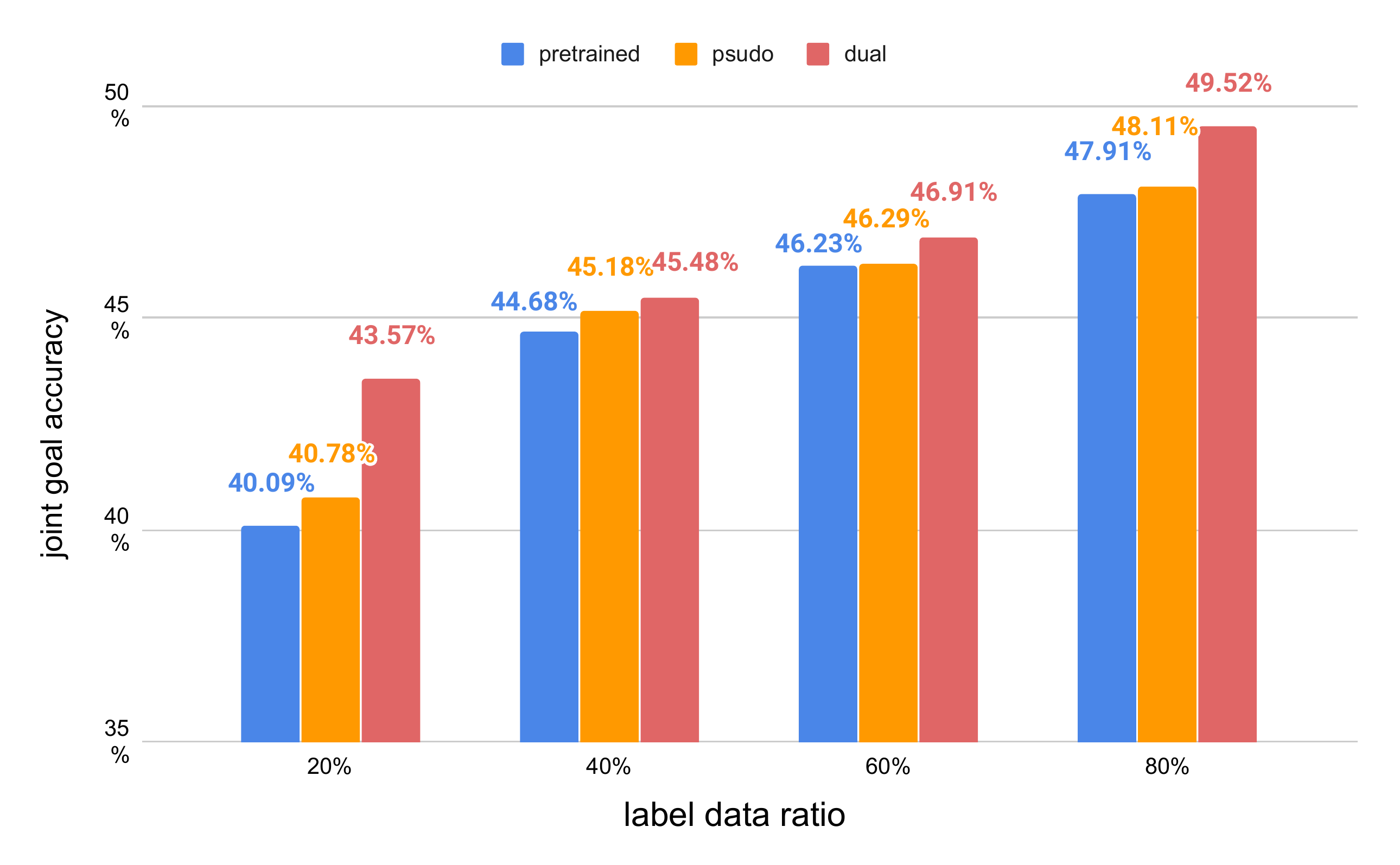}
    \caption{The joint goal accuracy with unlabeled data.}
    \label{fig:dual_exp}
\end{figure}

\paragraph{The performance of dual learning:}
In order to validate the effectiveness of our proposed dual learning framework, we randomly sample parts of training dataset as the small training set. The rest data is regarded as unlabeled data. In this experiment, we randomly sample 20\%, 40\%, 60\% and 80\% labeled data. As shown in Fig~\ref{fig:dual_exp}, we can see that the joint goal accuracy improves as the labeled data increases. It indicates that the scale of the annotated data is a big challenge for the multi-domain DST task. In this work, we propose a dual learning framework for DST to help to improve the performance of the tracker with the unlabeled data. 

When the training data is starved, the dual learning method can improve the performance of the pretrained tracker by efficiently using the unlabeled data. Compared with pseudo labeling method, our proposed dual learning method is able to treat the external knowledge (two kinds of language models: coarse state language model and user utterance language model) as reward function to feedback to the tracker and improve the performance. Especially, when the labeled data is extremely limited that only has 20\% sampled data, the dual learning method achieves a larger performance gain than the pseudo labeling method. As shown in Fig~\ref{fig:dual_exp}, we can see that the pseudo labeling method only gets less improvement. As we introduce in Section~\ref{sec:dataset}, the multi-domain DST task in MultiWOZ 2.1 dataset is more complex than single-domain DST task. The positive influence of the pseudo labeled data for pretrained tracker is limited. 

When the training data is fully used, the dual learning method can be still used to fine-tune the pretrained tracker. During the dual learning process, all the training data can be regarded as the unlabeled data. As shown in Table~\ref{tab:result}, the dual-DST can get further improvement from the fully pretrained tracker.

\section{Related Work}
\paragraph{Multi-Domain DST:} With the release of MultiWOZ dataset~\cite{budzianowski2018multiwoz}, one of the largest task-oriented dialogue datasets, many advanced dialogue state tracking methods for multi-domain task have been proposed. The previously proposed multi-domain state tracking approaches can be divided into two categories: classification~\cite{eric2019multiwoz} and generation~\cite{wu-2019-transferable}. The classification methods usually require that all the possible slot values are given by ontology. However, in real dialogue scenarios, some slot values cannot be enumerated. To alleviate this problem, the generative methods have been proposed, where the slot values are directly generated from the dialogue history. Like the classification methods, most of the generative methods generate slot value one by one, until all the slots on different domains have been visited. The methods that predict the slot values independently can not be used in dual learning framework. In this work, we redefine the dialogue state as a structured representation. We further propose a coarse-to-fine tracking method to directly generate the structured dialogue state.

\paragraph{Dual learning:}
Dual learning method is first proposed to improve neural machine translation (NMT)~\cite{he2016dual}. In NMT task, the primal task and the dual task are symmetric, while not in DST task. We design a state tracking model and an utterance generation model under the dual learning framework of DST. The idea of dual learning has been applied into various tasks, such as Question Answer~\cite{tang2017question}/ Generation~\cite{tang2018learning},  Image-to-Image Translatio~\cite{yi2017dualgan}, Open-domain Information Extraction/Narration~\cite{sun2018logician} and Semantic Parsing~\cite{cao2019semantic}. To the best of our knowledge, we are the first to introduce the dual learning in dialogue state tracking.

\section{Conclusion}
In this work, we first reformulate the dialogue state tracking task as a sequence generation task. Then we adopt a coarse-to-fine decoding method to directly generate the structured state sequence. The proposed coarse-to-fine tracker achieves the best performance among BERT-free methods. The main contribution of this work lies on building a dual learning framework for multi-domain DST task. The experimental results indicate that our proposed dual learning method can efficiently improve the pretrained tracker with unlabeled data. In future work, we will further improve the state tracking model and dual utterance generation model using pretrained models, e.g. BERT.

\newpage
\bibliographystyle{named}
\bibliography{ijcai20}

\begin{thebibliography}{}

\bibitem[\protect\citeauthoryear{Budzianowski \bgroup \em et al.\egroup
  }{2018}]{budzianowski2018multiwoz}
Pawe{\l} Budzianowski, Tsung-Hsien Wen, Bo-Hsiang Tseng, I{\~n}igo Casanueva,
  Stefan Ultes, Osman Ramadan, and Milica Gasic.
\newblock Multiwoz-a large-scale multi-domain wizard-of-oz dataset for
  task-oriented dialogue modelling.
\newblock In {\em Proceedings of the 2018 Conference on Empirical Methods in
  Natural Language Processing}, pages 5016--5026, 2018.

\bibitem[\protect\citeauthoryear{Cao \bgroup \em et al.\egroup
  }{2019}]{cao2019semantic}
Ruisheng Cao, Su~Zhu, Chen Liu, Jieyu Li, and Kai Yu.
\newblock Semantic parsing with dual learning.
\newblock In {\em Proceedings of the 57th Annual Meeting of the Association for
  Computational Linguistics}, pages 51--64, 2019.

\bibitem[\protect\citeauthoryear{Devlin \bgroup \em et al.\egroup
  }{2018}]{devlin2018bert}
Jacob Devlin, Ming-Wei Chang, Kenton Lee, and Kristina Toutanova.
\newblock Bert: Pre-training of deep bidirectional transformers for language
  understanding.
\newblock {\em arXiv preprint arXiv:1810.04805}, 2018.

\bibitem[\protect\citeauthoryear{Dong and Lapata}{2018}]{dong2018coarse}
Li~Dong and Mirella Lapata.
\newblock Coarse-to-fine decoding for neural semantic parsing.
\newblock In {\em Proceedings of the 56th Annual Meeting of the Association for
  Computational Linguistics (Volume 1: Long Papers)}, pages 731--742, 2018.

\bibitem[\protect\citeauthoryear{Eric \bgroup \em et al.\egroup
  }{2019}]{eric2019multiwoz}
Mihail Eric, Rahul Goel, Shachi Paul, Abhishek Sethi, Sanchit Agarwal, Shuyag
  Gao, and Dilek Hakkani-Tur.
\newblock Multiwoz 2.1: Multi-domain dialogue state corrections and state
  tracking baselines.
\newblock {\em arXiv preprint arXiv:1907.01669}, 2019.

\bibitem[\protect\citeauthoryear{Gao \bgroup \em et al.\egroup
  }{2019}]{gao2019dialog}
Shuyang Gao, Abhishek Sethi, Sanchit Agarwal, Tagyoung Chung, Dilek
  Hakkani-Tur, and Amazon~Alexa AI.
\newblock Dialog state tracking: A neural reading comprehension approach.
\newblock In {\em 20th Annual Meeting of the Special Interest Group on
  Discourse and Dialogue}, page 264, 2019.

\bibitem[\protect\citeauthoryear{Goel \bgroup \em et al.\egroup
  }{2019}]{goel2019hyst}
Rahul Goel, Shachi Paul, and Dilek Hakkani-T{\"u}r.
\newblock Hyst: A hybrid approach for flexible and accurate dialogue state
  tracking.
\newblock {\em Proc. Interspeech 2019}, pages 1458--1462, 2019.

\bibitem[\protect\citeauthoryear{Hashimoto \bgroup \em et al.\egroup
  }{2017}]{hashimoto2017joint}
Kazuma Hashimoto, Yoshimasa Tsuruoka, Richard Socher, et~al.
\newblock A joint many-task model: Growing a neural network for multiple nlp
  tasks.
\newblock In {\em Proceedings of the 2017 Conference on Empirical Methods in
  Natural Language Processing}, pages 1923--1933, 2017.

\bibitem[\protect\citeauthoryear{He \bgroup \em et al.\egroup
  }{2016}]{he2016dual}
Di~He, Yingce Xia, Tao Qin, Liwei Wang, Nenghai Yu, Tie-Yan Liu, and Wei-Ying
  Ma.
\newblock Dual learning for machine translation.
\newblock In {\em Advances in neural information processing systems}, pages
  820--828, 2016.

\bibitem[\protect\citeauthoryear{Kim \bgroup \em et al.\egroup
  }{2019}]{kim2019efficient}
Sungdong Kim, Sohee Yang, Gyuwan Kim, and Sang-Woo Lee.
\newblock Efficient dialogue state tracking by selectively overwriting memory.
\newblock {\em arXiv preprint arXiv:1911.03906}, 2019.

\bibitem[\protect\citeauthoryear{Kingma and Ba}{2014}]{kingma2014adam}
Diederik~P Kingma and Jimmy Ba.
\newblock Adam: A method for stochastic optimization.
\newblock {\em arXiv preprint arXiv:1412.6980}, 2014.

\bibitem[\protect\citeauthoryear{Luong \bgroup \em et al.\egroup
  }{2015}]{luong2015effective}
Minh-Thang Luong, Hieu Pham, and Christopher~D Manning.
\newblock Effective approaches to attention-based neural machine translation.
\newblock In {\em Proceedings of the 2015 Conference on Empirical Methods in
  Natural Language Processing}, pages 1412--1421, 2015.

\bibitem[\protect\citeauthoryear{Mikolov \bgroup \em et al.\egroup
  }{2010}]{mikolov2010recurrent}
Tom{\'a}{\v{s}} Mikolov, Martin Karafi{\'a}t, Luk{\'a}{\v{s}} Burget, Jan
  {\v{C}}ernock{\`y}, and Sanjeev Khudanpur.
\newblock Recurrent neural network based language model.
\newblock In {\em Eleventh annual conference of the international speech
  communication association}, 2010.

\bibitem[\protect\citeauthoryear{Pennington \bgroup \em et al.\egroup
  }{2014}]{pennington2014glove}
Jeffrey Pennington, Richard Socher, and Christopher Manning.
\newblock Glove: Global vectors for word representation.
\newblock In {\em Proceedings of the 2014 conference on empirical methods in
  natural language processing (EMNLP)}, pages 1532--1543, 2014.

\bibitem[\protect\citeauthoryear{Sun \bgroup \em et al.\egroup
  }{2018}]{sun2018logician}
Mingming Sun, Xu~Li, and Ping Li.
\newblock Logician and orator: Learning from the duality between language and
  knowledge in open domain.
\newblock In {\em Proceedings of the 2018 Conference on Empirical Methods in
  Natural Language Processing}, pages 2119--2130, 2018.

\bibitem[\protect\citeauthoryear{Sutton \bgroup \em et al.\egroup
  }{2000}]{sutton2000policy}
Richard~S Sutton, David~A McAllester, Satinder~P Singh, and Yishay Mansour.
\newblock Policy gradient methods for reinforcement learning with function
  approximation.
\newblock In {\em Advances in neural information processing systems}, pages
  1057--1063, 2000.

\bibitem[\protect\citeauthoryear{Tang \bgroup \em et al.\egroup
  }{2017}]{tang2017question}
Duyu Tang, Nan Duan, Tao Qin, Zhao Yan, and Ming Zhou.
\newblock Question answering and question generation as dual tasks.
\newblock {\em arXiv preprint arXiv:1706.02027}, 2017.

\bibitem[\protect\citeauthoryear{Tang \bgroup \em et al.\egroup
  }{2018}]{tang2018learning}
Duyu Tang, Nan Duan, Zhao Yan, Zhirui Zhang, Yibo Sun, Shujie Liu, Yuanhua Lv,
  and Ming Zhou.
\newblock Learning to collaborate for question answering and asking.
\newblock In {\em Proceedings of the 2018 Conference of the North American
  Chapter of the Association for Computational Linguistics: Human Language
  Technologies, Volume 1 (Long Papers)}, pages 1564--1574, 2018.

\bibitem[\protect\citeauthoryear{Vaswani \bgroup \em et al.\egroup
  }{2017}]{vaswani2017attention}
Ashish Vaswani, Noam Shazeer, Niki Parmar, Jakob Uszkoreit, Llion Jones,
  Aidan~N Gomez, {\L}ukasz Kaiser, and Illia Polosukhin.
\newblock Attention is all you need.
\newblock In {\em Advances in neural information processing systems}, pages
  5998--6008, 2017.

\bibitem[\protect\citeauthoryear{Wu \bgroup \em et al.\egroup
  }{2019}]{wu-2019-transferable}
Chien-Sheng Wu, Andrea Madotto, Ehsan Hosseini-Asl, Caiming Xiong, Richard
  Socher, and Pascale Fung.
\newblock Transferable multi-domain state generator for task-oriented dialogue
  systems.
\newblock In {\em Proceedings of the 57th Annual Meeting of the Association for
  Computational Linguistics}, pages 808--819, Florence, Italy, July 2019.
  Association for Computational Linguistics.

\bibitem[\protect\citeauthoryear{Yi \bgroup \em et al.\egroup
  }{2017}]{yi2017dualgan}
Zili Yi, Hao Zhang, Ping Tan, and Minglun Gong.
\newblock Dualgan: Unsupervised dual learning for image-to-image translation.
\newblock In {\em Proceedings of the IEEE international conference on computer
  vision}, pages 2849--2857, 2017.

\bibitem[\protect\citeauthoryear{Zhang \bgroup \em et al.\egroup
  }{2019}]{zhang2019find}
Jian-Guo Zhang, Kazuma Hashimoto, Chien-Sheng Wu, Yao Wan, Philip~S Yu, Richard
  Socher, and Caiming Xiong.
\newblock Find or classify? dual strategy for slot-value predictions on
  multi-domain dialog state tracking.
\newblock {\em arXiv preprint arXiv:1910.03544}, 2019.

\end{thebibliography}

\end{document}